%% file: ijcai24.tex
\documentclass{article}
\usepackage{ijcai24}

\usepackage{times}
\usepackage{soul}
\usepackage{url}
\usepackage[hidelinks]{hyperref}
\usepackage[utf8]{inputenc}
\usepackage[small]{caption}
\usepackage{graphicx}
\usepackage{amsmath}
\usepackage{amsthm}
\usepackage{booktabs}
\usepackage{algorithm}
\usepackage{algorithmic}
\usepackage[switch]{lineno}

\usepackage{amsfonts}
\usepackage{stfloats}
\usepackage{float}
\usepackage{caption}
\usepackage{marvosym}

\title{BoostDream: Efficient Refining for High-Quality Text-to-3D Generation from Multi-View Diffusion}

\author{
Yonghao Yu$^1$\footnote{Corresponding author}
\and
Shunan Zhu$^1$\and
Huai Qin$^1$\And
Haorui Li$^2$\
\affiliations
$^1$Waseda University\\
$^2$ Southeast University\\
\emails
yuyonghao@suou.waseda.jp,
\{shunan-zhu, mizuki\_qin\}@ruri.waseda.jp,
lihaorui.lhr@alibaba-inc.com,
}

\begin{document}

\maketitle

\begin{abstract}

Witnessing the evolution of text-to-image diffusion models, significant strides have been made in text-to-3D generation. Currently, two primary paradigms dominate the field of text-to-3D: the feed-forward generation solutions, capable of swiftly producing 3D assets but often yielding coarse results, and the Score Distillation Sampling (SDS) based solutions, known for generating high-fidelity 3D assets albeit at a slower pace. The synergistic integration of these methods holds substantial promise for advancing 3D generation techniques.
In this paper, we present BoostDream, a highly efficient plug-and-play 3D refining method designed to transform coarse 3D assets into high-quality. The BoostDream framework comprises three distinct processes: (1) We introduce 3D model distillation that fits differentiable representations from the 3D assets obtained through feed-forward generation.
(2) A novel multi-view SDS loss is designed, which utilizes a multi-view aware 2D diffusion model to refine the 3D assets.
(3) We propose to use prompt and multi-view consistent normal maps as guidance in refinement.
Our extensive experiment is conducted on different differentiable 3D representations, revealing that BoostDream excels in generating high-quality 3D assets rapidly, overcoming the Janus problem compared to conventional SDS-based methods. This breakthrough signifies a substantial advancement in both the efficiency and quality of 3D generation processes.

\end{abstract}

 \begin{figure*}[ht]
     \centering
    \includegraphics[width= 0.95\textwidth]{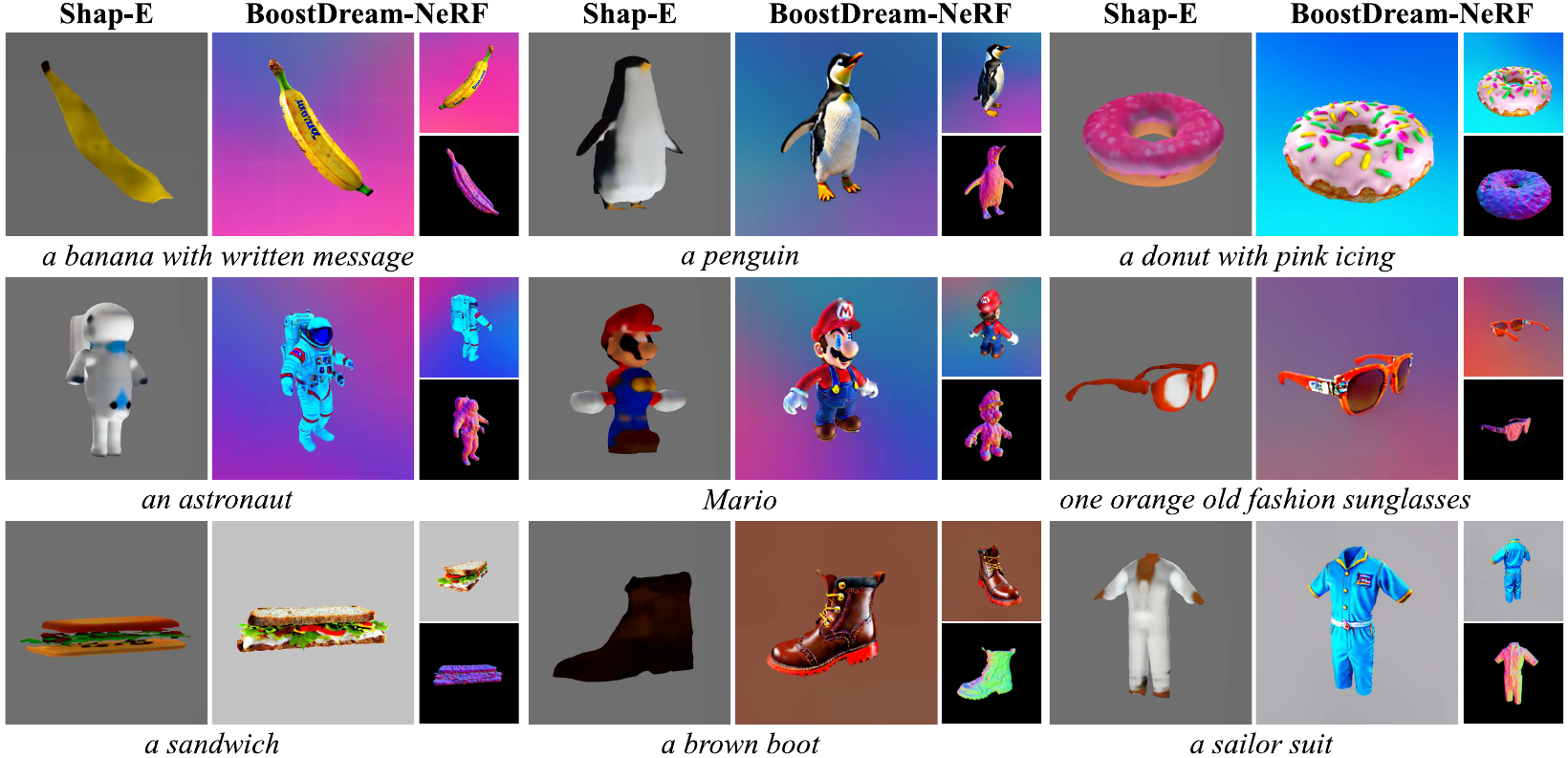}
    \caption{\textbf{Comparison of 3D Generation Results of baseline and \emph{BoostDream}.} Provided with a coarse 3D asset and text prompt pair, BoostDream can refine it into a high-quality 3D asset efficiently. In each set of images, the image on the left is the coarse 3D asset generated by Shap-E \protect\cite{jun2023shape}, and the three images on the right are our refined 3D asset. }
    \label{fig:figure1}
\end{figure*}

\section{Introduction}
The significance of 3D assets has grown remarkably due to the expansion of virtual reality and the gaming industry. Creating these assets, however, involves considerable labor costs. Recently, the popularity of methods based on differentiable rendering \cite{mildenhall2021nerf,shen2021dmtet,wang2021neus,kerbl3Dgaussians} and the explosion of text-to-image models \cite{rombach2022high,zhang2023adding} provide a new line of methods for generating 3D assets. 


Plenty of research \cite{nichol2022pointe,jun2023shape,poole2022dreamfusion,wang2023prolificdreamer,lin2023magic3d,chen2023fantasia3d} in the text-to-3D domain aim to lower 3D asset creation costs, demonstrating the efficacy of leveraging 2D generation models for 3D tasks. Currently, there are two primary strategies in the field of text-to-3D generation. The first, exemplified by Point-E and Shap-E \cite{nichol2022pointe,jun2023shape}, employs a feed-forward approach, using 3D datasets or latent representation mapped by the 3D datasets to train diffusion models. This enables quick, text-prompt-based 3D asset generation. However, these methods grapple with the limited variety and scope of 3D datasets.

In response to the constraints posed by limited 3D datasets, the SDS-based optimization method was developed. Models like DreamFusion and Magic3D \cite{poole2022dreamfusion,lin2023magic3d} exemplify this approach. 
They employ an SDS loss function \cite{poole2022dreamfusion}, enabling text-to-image diffusion models to train 3D assets predominantly represented by methods based on differentiable rendering, such as NeRF \cite{mildenhall2021nerf}.
This approach allows for the generation of high-quality 3D assets, leveraging the strong text comprehension and 2D generation abilities of diffusion models. However, this method is hindered by slow generation speeds, as each text prompt requires training a 3D representation from scratch and suffering from the Janus (multi-head) problem.
How to efficiently generate high-quality 3D assets remains an urgent problem to be solved.

With this in mind, we propose a generation method suitable for various 3D representations, utilizing the rapid generation of the feed-forward method and the high-quality generation characteristics of the SDS-based method.
Our method does not just simply combine the feed-forward approach with the SDS-based method. Specifically, we have developed an innovative three-stage method named BoostDream. In the first stage, we designed a rapid initialization technique that transforms explicitly represented 3D assets generated by the feed-forward method into 3D representations based on differentiable rendering. In the second stage, we introduced a multi-view rendering system and a multi-view SDS under the control of the original 3D input condition. In the third stage, we solely rely on self-supervision to achieve refined generation results. Extensive experiments are conducted, including refinement and comparison experiments. As shown in Figure \ref{fig:figure1}, our method can efficiently refine the coarse 3D for high-quality results.  In summary, the contributions of this paper include the following:
\begin{itemize}
    \item We propose a novel method that integrates the advances in feed-forward and SDS-based methods, enabling efficient and high-quality refinement of 3D assets.
    \item We innovatively propose the multi-view SDS with the multi-view render system to refine 3D assets under multi-view conditions and address the Janus problem.
    \item Our BoostDream can generate high-quality 3D assets based on a variety of 3D differentiable representations, showcasing strong generalizability.
\end{itemize}

\section{Related Work}
In this section, we discuss the related work on text-to-3D generation using diffusion models. One mainstream approach is the feed-forward method using 3D datasets for training, and another is SDS-based optimization. In addition, the multi-view perceptual diffusion method provides ideas for solving the Janus problem and has proved effective.

\subsection{Feed-Forward Generation Method}
Feed-forward generation method tries to directly utilize the diffusion models in text-to-3D tasks, which trained diffusion models with 3D assets as datasets \cite{liu2023meshdiffusion,liu2023one2345,sanghi2023clip,nichol2022pointe,zhou20213d} or perform diffusion with a latent representation of 3D assets \cite{gupta20233dgen,ntavelis2023autodecoding,chen2023single,zeng2022lion}. Shap-E \cite{jun2023shape} is a typical example of this type of approach. It trains a 3D encoder to transform 3D assets to a set of parameters of an implicit function and then trains a diffusion model on these parameters. Compared to the methods using frozen 2D diffusion models, these approaches can generate 3D assets at a very fast speed due to their feed-forward nature. However, these methods were trained with the 3D datasets, while the 3D dataset available today is relatively small and low-quality. Consider the dataset used in 2D image generation tasks, Laion5B \cite{schuhmann2022laion5b} contains more than 5 billion image-text pairs while the largest 3D dataset available, Objaverse-XL \cite{deitke2023objaverse} can only carry 10 million 3D assets with worse quality captions. This causes the 3D generation results to be less sharp, have lower fidelity, and be unable to produce assets with complex semantics.

\subsection{SDS-Based Optimization Generation Method}

Inspired by the great success of using diffusion models in text-to-image tasks, one mainstream of works focuses on using 2D diffusion models along with differentiable 3D representing techniques like NeRF \cite{mildenhall2021nerf} to facilitate 3D generation. Dreamfields \cite{jain2022dreamfields} used pre-trained CLIP models to optimize NeRF, hoping to get a visually real result. But it takes much time and computing resources to generate. DreamFusion \cite{poole2022dreamfusion} proposes SDS loss to optimize NeRF representation. SJC \cite{wang2023score} is a concurrent effort of Dreamfusion, with the difference that SJC applies the chain rule to the 2D score. Research on providing supervision of novel views via SDS loss is becoming a trend \cite{lin2023magic3d,chen2023fantasia3d,metzer2023latent,liu2023zero,qian2023magic123,melas2023realfusion,lorraine2023att3d}.  Among them, ProlificDreamer \cite{wang2023prolificdreamer} models the 3D parameters as variable, proposes variable fraction distillation (VSD) to address the shortcomings in SDS. These works indicate the capability to produce high-quality outputs enriched with detailed semantic information derived from 2D diffusion models. However, these SDS-based methods are impeded by prolonged optimization periods and the occurrence of Janus problems.

\subsection{Multi-View in 3D Generation}
The multi-view aware diffusion method in image generation has made great progress recently. MVDiffusion \cite{tang2023mvdiffusion} is capable of generating consistent multi-view images from text prompts given pixel-to-pixel correspondences with global awareness. The success of such a method also draws attention to 3D generation tasks. EfficientDreamer \cite{zhao2023efficientdreamer} used a 2D diffusion model that can generate an image consisting of four orthogonal-view sub-images and use it in the 3D generation. It has been proven to be efficient in improving the quality of 3D content and alleviating the Janus problem. 
MVDream \cite{shi2023mvdream} further discusses the situation of building a multi-view diffusion model from both 2D and 3D data. It can benefit from the generalizability of the 2D diffusion model while keeping the consistency of 3D rendering, solving the Janus problem and content drift problem across different views. However, after training by the 3D datasets, the realistic texture information that the 2D diffusion model contained can be degraded. In our work, a multi-view rendering method is designed and a multi-view SDS loss function is formulated to guide the differentiable 3D representation, mitigating the Janus problem while obtaining high-quality generated results.

\newcommand{\model}{\epsilon}
\newcommand{\decoder}{\mathcal{D}}
\newcommand{\encoder}{\mathcal{E}}
\newcommand{\cond}{y}
\newcommand{\conditioner}{\tau_\theta}
 \section{BoostDream}

 \begin{figure*}[htp]
     \centering
    \includegraphics[width=0.97\textwidth]{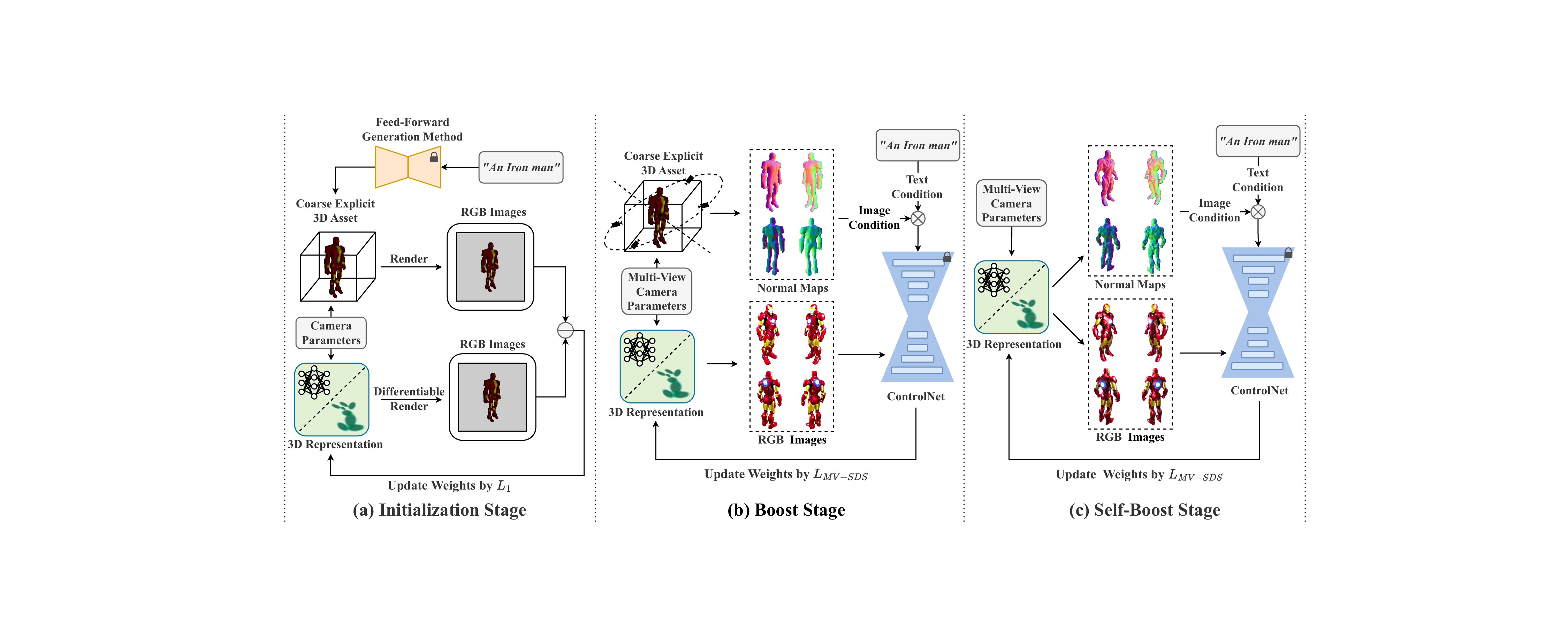}
    \caption{\textbf{Overview of the proposed \emph{BoostDream}}. BoostDream is a three-stage framework for refining a coarse 3D asset into a high-quality 3D asset. In the initialization stage, we use the feed-forward generation method to get a coarse 3D asset and fit it into differentiable 3D representations to make it trainable. The boost stage is guided by the multi-view normal maps of the coarse 3D asset to ensure stability from the beginning of the refining stage, and the self-boost stage is guided by its own multi-view normal maps to generate 3D assets with more detail and higher quality.}
    \label{fig:figure2}
\end{figure*}

In this section, we propose our BoostDream model. We initially present a brief overview of the background of text-to-3D generation. Subsequently, we introduce our multi-view refining approach, detailing its design and implementation within the context of enhanced 3D asset generation by quick initialization based on differentiable rendering, multi-view render system, and multi-view SDS. Figure \ref{fig:figure2} illustrates the overview of our method.


\subsection{Background}

For text-to-image tasks, \cite{NEURIPS2020_4c5bcfec} has proposed a simplified training goal of Denoising Diffusion Probabilistic Models (DDPM). It is expressed as:
\begin{equation}
\begin{split}
&L_{\text {DDPM}}= \\ & \mathbb{E}_{t, \mathbf{x}_0, \boldsymbol{\epsilon}}\left[\left\|\boldsymbol{\epsilon}-\boldsymbol{\epsilon}_\theta\left(\sqrt{\bar{\alpha}_t} \mathbf{x}_0+\sqrt{1-\bar{\alpha}_t} \boldsymbol{\epsilon}, t\right)\right\|^2\right]
\end{split}
\end{equation}
Here the $x_t$ is latent at time step $t$, $\epsilon$ is the actual noise, $\epsilon_{\theta}$ represents the predicted noise, $\alpha_t = 1-\beta_t$ where $\beta_t$ is the noise added at time step $t$ in a diffusion model.



In text-to-3D tasks using text-prompt as the generation condition, DreamFusion \cite{poole2022dreamfusion,wang2023prolificdreamer} has proposed the SDS loss. For a differentiable 3D rendering represented as $g(\theta, c)$, where $\theta$ are its parameters and $c$ are the camera parameters. 
Let $q_t^\theta\left(\mathbf{x}_t \mid c\right)$ be the distribution from diffusing $\mathbf{x}_0=g(\theta, c)$ to moment $t$. The optimization goal of SDS is by optimizing $\theta$, make the marginal distribution of the diffusion process of $q_t^\theta\left(\mathbf{x}_t \mid c\right)$, approach the marginal distribution of a pre-trained text-to-image model $p\left(\mathbf{x}_t \mid y\right)$ where $y$ is text-prompt. The training goal of the SDS loss method can be represented as:
\begin{equation}
\begin{split}
 & \min _\theta \mathcal{L}_{\mathrm{SDS}}  = \\ & \mathbb{E}_{t, c}\left[\left(\sigma_t / \alpha_t\right) w(t) D_{K L}\left(q_t^\theta\left(\mathbf{x}_t \mid c\right) \| p\left(\mathbf{x}_t \mid y\right)\right)\right]
\end{split}
\end{equation}

The SDS method estimates the gradient of the optimization object $\mathcal{L}_{\mathrm{SDS}}$ as follows, allowing the text-to-3D task to be carried out quickly and without losing too much precision:
\begin{equation}
\begin{split}
\nabla_\theta \mathcal{L}_{\mathrm{SDS}} = \mathbb{E}_{t, \epsilon, c}\left[w(t)\left(\epsilon_\theta\left(\mathbf{x}_t, t, y\right)-\epsilon\right) \frac{\partial g(\theta, c)}{\partial \theta}\right]
\label{eq:gradient}
\end{split}
\end{equation}

The idea of adding other control conditions in BoostDream is inspired by ControlNet \cite{controlnetmodel}, which is a text-to-image task solution allowing more input as task specific conditions like images. Its training goal can be written as:
 \begin{equation}
 \mathcal{L_{\text{Control}}} = \mathbb{E}_{x_0, t, c_t, c_f, \epsilon} \left[\| \epsilon - \epsilon_{\theta}(x_t, t, c_t, c_f )\|_{2}^{2}\right]
 \end{equation}
In this formulation, the terms $c_t$ and $c_f$ are task-specific conditions applied in ControlNet, like text-prompt and images.

\subsection{3D Representation Initialization}

Our method is applicable to a variety of 3D asset generation approaches based on differentiable rendering. In order to achieve this, we propose a generalized model distillation approach to fit a coarse 3D asset into a randomly initialized differential representation.

We define the coarse explicit 3D assets as $a$ and the parameters of differentiable 3D representation as $\theta$. In every iteration, the randomly sampled camera parameters $c$ are passed into the renderer $g$ to get the rendered images for these two 3D assets separately at the same view. With these pairs of images, we use $L_1$ loss function to make the randomly initialized 3D representation look as much like the coarse explicit 3D asset as possible in a very short time. The loss function is defined as:
\begin{equation}
L_1=|g(a,c)-g(\theta,c)|
\end{equation}


\subsection{Multi-View Render System}\label{sec:render}
To alleviate the Janus problem, we designed a new multi-view rendering method. As shown in the two boost stages of Figure \ref{fig:figure2}, we initialize the default viewing direction in accordance with the input 3D model whose origin coincides with the origin of the coordinate system. At each iteration, we randomly sampled a camera position $p_0$ in spherical coordinates, consisting of elevation angle $\phi_{cam}\in[-10^\circ,70^\circ]$, azimuth angle $\theta_{cam}\in[0^\circ,360^\circ]$, field of view  $f\in[0^\circ,180^\circ]$ and camera distance $d_{cam}$. 
We further created a rotation axis, a random vector passing through the origin denoted as $\mathbf{a} = (a_x, a_y, a_z)$, a rotation angle denoted as $\alpha$. To get multi-view camera position, a rotation matrix $\mathbf{R}$ is defined as:
\begin{equation}
\mathbf{R}(\mathbf{a}, \alpha) = \mathbf{I} + (\sin \alpha) \mathbf{K} + (1 - \cos \alpha) \mathbf{K}^2 \
\end{equation}
Where \( \mathbf{I} \) is the identity matrix, and \( \mathbf{K} \) is the skew-symmetric matrix generated from the rotation axis \( \mathbf{a} = (a_x, a_y, a_z) \), defined as:
\begin{equation}
\mathbf{K} = \begin{pmatrix} 0 & -a_z & a_y \\ a_z & 0 & -a_x \\ -a_y & a_x & 0 \end{pmatrix} 
\end{equation}

In our experimental settings, we define rotation angle $\alpha=90^\circ$, so the camera at position ${p}_0$ will rotate around axis $\mathbf{a}$ and get four camera positions. In each rotation iteration $i$, the camera position ${p}_i$ after rotation follows:
\begin{equation}
{p}_i = \mathbf{R}(\mathbf{a}, i \times \alpha) {p}_0
\end{equation}

We use separate renderers for each task in different 3D representations. For a renderer $g$, we use the above four camera positions $p_i$ along with other parameters to get camera parameters $c_i$. With a total of four sub-images generated from different camera parameters $c_i$ and parameters of a 3D representation $\theta$, stitch these four images together to create one large $2\times2$ composite image ${G}$. We also capture the normal maps rendered from 3D representation and follow a similar routine to get a large $2\times2$ normal map ${N}$.
\begin{equation}
\label{eq.G.final}
G = \begin{pmatrix} g(\mathbf{\theta}, c_0) &g(\mathbf{\theta}, c_1)\\g(\mathbf{\theta}, c_2)& g(\mathbf{\theta}, c_3) \end{pmatrix} 
\end{equation}

\subsection{Multi-View SDS}

The multi-view SDS loss function is designed to direct differentiable rendering models, utilizing both text prompt and multi-view conditions generated by the camera parameters we proposed in section \ref{sec:render} as the control guidance.
We use multi-view normal maps here for multi-view conditions, as they provide a detailed depiction of surface normals, enhancing the capture of finer geometric and textural details. This enhancement is particularly beneficial for accurately preserving complex features like hair or surface irregularities, which are essential for lifelike 3D renderings.

The heart of this loss function is a new method of estimating noise guided by dual task-specific conditions, like normal maps and the text prompt. This dual-conditioned approach ensures that the generated content remains faithful to the surface details outlined by the normal maps while also aligning with the context provided by the text prompt. The noise estimation is formulated as:
\begin{equation}
\begin{split}
\hat{\epsilon}_{\phi}(x_{t}; t, y, N) = & \ \epsilon_{\phi}(x_{t}; t, y, \lambda * N) \\
& + s * \left(\epsilon_{\phi}(x_{t}; t, y, \lambda * N) - \epsilon_{\phi}(x_{t}; t)\right)
\end{split}
\end{equation}
Here, \( \epsilon_{\phi} \) denotes the noise predicted by the diffusion model, \( \lambda \in [0,1] \)  balances the normal map conditioned and unconditioned noise predictions \cite{chen2023control3d}, and \( s \) is the scale of classifier-free guidance \cite{ho2022classifierfree}. \( N \) represents the multi-view normal maps captured following the rule we proposed in section \ref{sec:render}, for normal map renderer $g_N$, we use the same camera parameters used by composite image ${G}$ in Eq. \ref{eq.G.final}:
\begin{equation}
N = \begin{pmatrix} g_N(\mathbf{\theta}, c_0) &g_N(\mathbf{\theta}, c_1)\\g_N(\mathbf{\theta}, c_2)& g_N(\mathbf{\theta}, c_3) \end{pmatrix} 
\end{equation}

Like Eq. \ref{eq:gradient}, the gradient of the multi-view SDS loss concerning the differentiable representations' parameters is defined as:
\begin{equation}
\begin{split}
\nabla_{\theta}\mathcal{L}_{MV-SDS} = \mathbb{E}_{t, \epsilon}\left[w(t)(\hat{\epsilon}_{\phi}(x_{t}; t, y, N) - \epsilon)\frac{\partial G}{\partial \theta}\right]
\end{split}
\end{equation}

Furthermore, We enhance the loss with orientation loss $\mathcal{L}_{\mathrm{orient}}$ and opacity loss $\mathcal{L}_{\mathrm{opacity}}$ proposed at \cite{poole2022dreamfusion}, when implement with NeRF. The final loss function combines these components to ensure accurate surface detail representation, correct surface orientations, and optimal opacity:
\begin{equation}
\mathcal{L}_{\text{total}} = \mathcal{L}_{MV-SDS}(\phi, x) + \alpha \mathcal{L}_{\mathrm{orient}} + \beta \mathcal{L}_{\mathrm{opacity}}
\end{equation}
where \( \alpha \) and \( \beta \) are the weights for the orientation and opacity losses, respectively. This comprehensive loss function fosters the generation of 3D content that is both semantically and geometrically accurate, exhibiting realistic physical properties.

\begin{figure*}[t]
    \centering
    \includegraphics[width=0.98\textwidth]{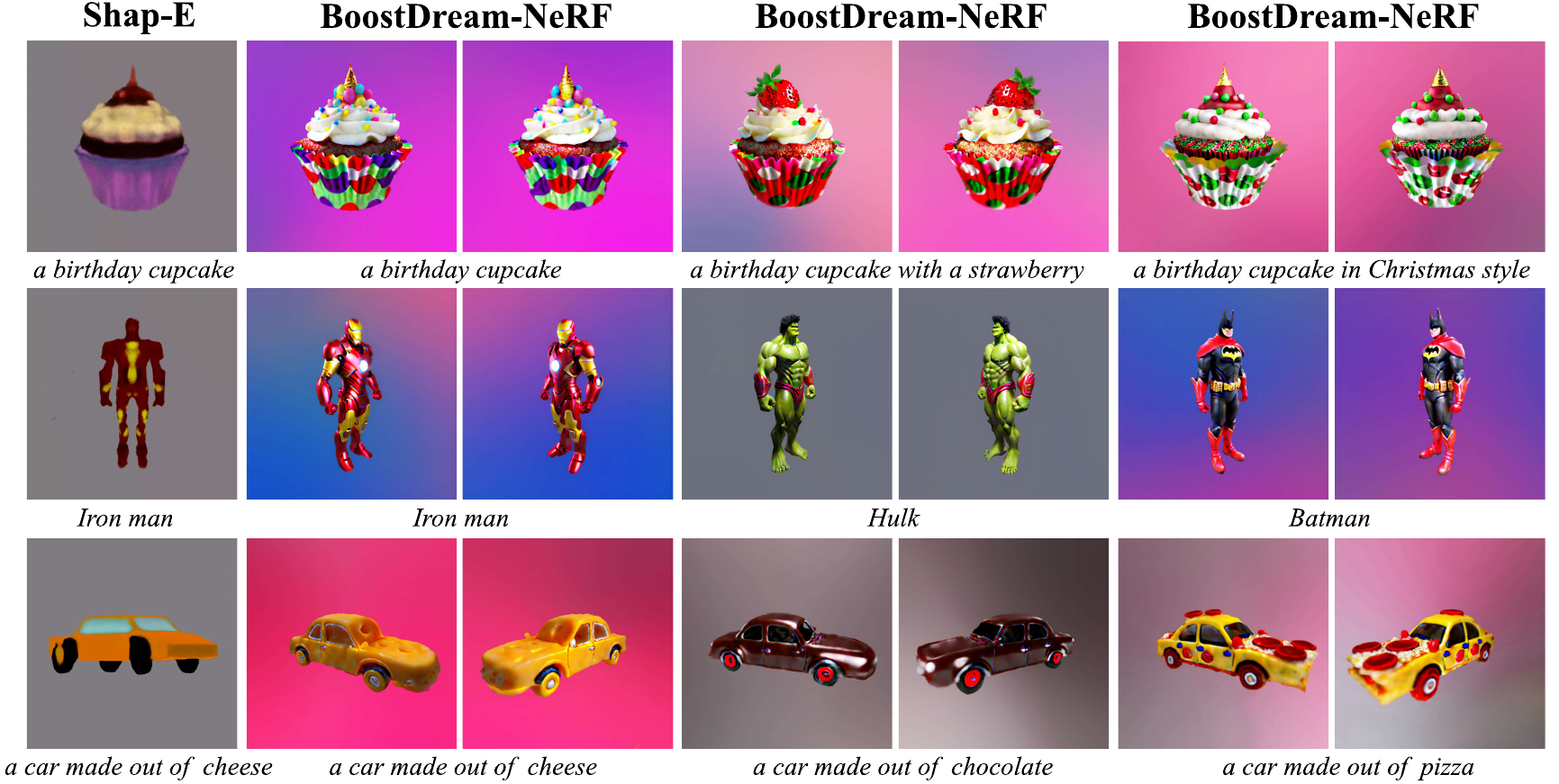}
    \caption{The first column is the Shap-E \protect\cite{jun2023shape} results and the remaining column is the refined results of our method. The results show that BoostDream can refine and edit 3D assets according to different prompts based on input 3D assets.}
    \label{fig:figure3}
\end{figure*}

\section{Experiments}
In this section, we present a series of experiments designed to evaluate the capabilities of our BoostDream method. In Section \ref{sec:implementation}, we detail the implementation specifics. Section \ref{sec:Refinement} discusses the results of refinement experiments and explores its performance under various text prompts. In Section \ref{sec:Comparison}, we compare our method against existing techniques in terms of performance and speed. We conduct some ablation studies to illustrate the effect of each stage and condition setting of our method in Section \ref{sec:Ablation}. We also defined a user study to quantify the model performance which is shown in Section \ref{sec:User}. These experiments are crucial in demonstrating the effectiveness of BoostDream in refining and editing 3D assets and the generalisability when applied to different 3D representations.

\subsection{Implementation Details}\label{sec:implementation}

All the experiments in this paper are conducted on a single NVIDIA V100 GPU with 32GB VRAM. We use Shap-E \cite{jun2023shape} from official implementation \cite{ShapE2023} to generate coarse 3D assets and fit them into differentiable 3D representations in the first 200 iterations. After initialization, we set 4800 iterations for the refining process, of which the first 1800 iterations are under the guidance of the coarse 3D assets, and the subsequent 3000 iterations are guided by the differentiable rendering result itself. We use ControlNet 1.1 with Stable Diffusion 1.5 \cite{controlnetmodel} as the diffusion model. In the comparison experiment, we carried out experiments on different differentiable 3D representations, including NeRF \cite{mildenhall2021nerf}, DMTet \cite{shen2021dmtet}, and 3D Gaussian Splatting \cite{kerbl3Dgaussians}. Please refer to the Appendix \cite{Boost2024} for more details and results.

\begin{figure*}[htp]
    \centering
    \includegraphics[width=0.94\textwidth]{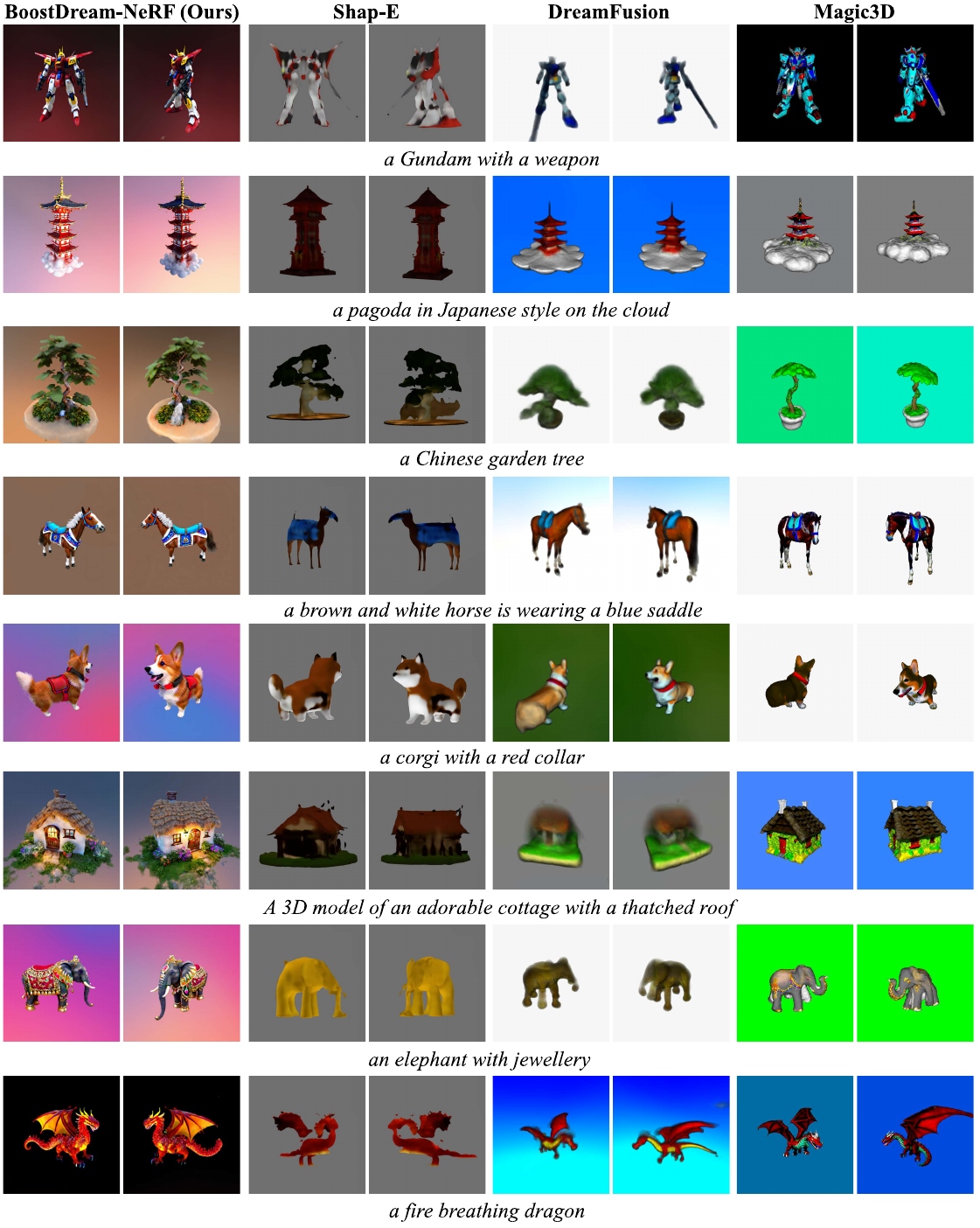}
    \caption{Comparision with Shap-E \protect\cite{jun2023shape}, DreamFusion \protect\cite{poole2022dreamfusion} and Magic3D \protect\cite{lin2023magic3d} for the same text-to-3D generation task. Our model has significantly stronger prompt relevancy and much better quality (best viewed by zooming in). See the results of our method on DMTet \protect\cite{shen2021dmtet} and 3D Gaussian Splatting \protect\cite{kerbl3Dgaussians} in the Appendix \protect\cite{Boost2024}}.
    \label{fig:figure4}
\end{figure*}

\subsection{Refinement Experiment} \label{sec:Refinement}
We designed a refinement experiment that shows the powerful refining capabilities of our BoostDream. We choose Shap-E as the feed-forward generation method to generate coarse 3D assets. The coarse 3D assets will be sent to our refining method with the same text prompt. We use NeRF implementation here as an example. The result is shown in Figure \ref{fig:figure1}. 
Our BoostDream can also refine coarse 3D assets with effective prompt-based editing. Experiments demonstrate its ability to refine assets using varied text prompts while retaining original features, as illustrated in Figure \ref{fig:figure3}. Notably, examples like ``Hulk" and ``Batman" preserve the original posture and red elements from the ``Iron Man" asset, showcasing a delicate balance between adaptation and preservation.

\subsection{Comparison Experiment}\label{sec:Comparison}

Our BoostDream is a refining method that combines the advantages of the feed-forward generation method and the SDS-based optimization generation method. To demonstrate the effectiveness of our methods, we selected methods from each of the above two categories for the comparison experiment. For the feed-forward generation method, we choose Shap-E, which has an official implementation \cite{ShapE2023}. The 3D assets generated by Shap-E are also used to initialize our BoostDream method. For the SDS-based optimization generation method, we choose DreamFusion and Magic3D, which are implemented by threestudio \cite{threestudio2023} because no official code is available now. It is also worth noticing there are some differences between the original papers and the threestudio implementation. In this implementation, DreamFusion and Magcic3D use DeepFloyd \cite{Deepfloyd} as the diffusion model, while the originals use Imagen \cite{saharia2022Imagen} and eDiff-I \cite{balaji2022ediffi}. We use the default parameter of these methods. 
The experiment result is shown in Figure \ref{fig:figure4}. Our method can generate higher-quality 3D assets compared to the above methods. In rows 1, 4, and 7, DreamFusion and Magic3D have arisen with the Janus problem, which means the Gundam has multiple arms, the horse has wrong legs, and the elephant has numerous heads, while our method can alleviate this problem due to the three-stage refine.


Also, compared to the two SDS-based optimization generation methods, the total generation time we need is reduced. Our method only requires 5000 iterations in total. As a comparison, the DreamFusion needs 15000 iterations to get a 3D asset, while the Magic3D needs 5000 iterations to train a coarse 3D asset and has a 3000 iteration refinement process.  The speed comparison results are shown in Table \ref{tab:tab1}. 
From the result, we can observe that apart from our high-quality generation result, BoostDream can rapidly finish optimization.

\begin{table}[h]
    \centering
    \begin{tabular}{lrr}
        \toprule
        Method  & Time (V100-sec) \\
        \midrule
        BoostDream-NeRF &  2038           \\
        BoostDream-DMTet &  2128           \\
        BoostDream-3D Gaussian Splatting &  1977           \\
        DreamFusion & 3519            \\
        Magic3D & 2355           \\
        \bottomrule
    \end{tabular}
    \caption{Speed comparison. Each time is the average training time of ten different text prompts. Note that DreamFusion and Magic3D are tested on threestudio implementation.}
    \label{tab:tab1}
\end{table}

\subsection{Ablation Study} \label{sec:Ablation}
\begin{figure}[ht]
    \centering
    \includegraphics[width = 8cm]{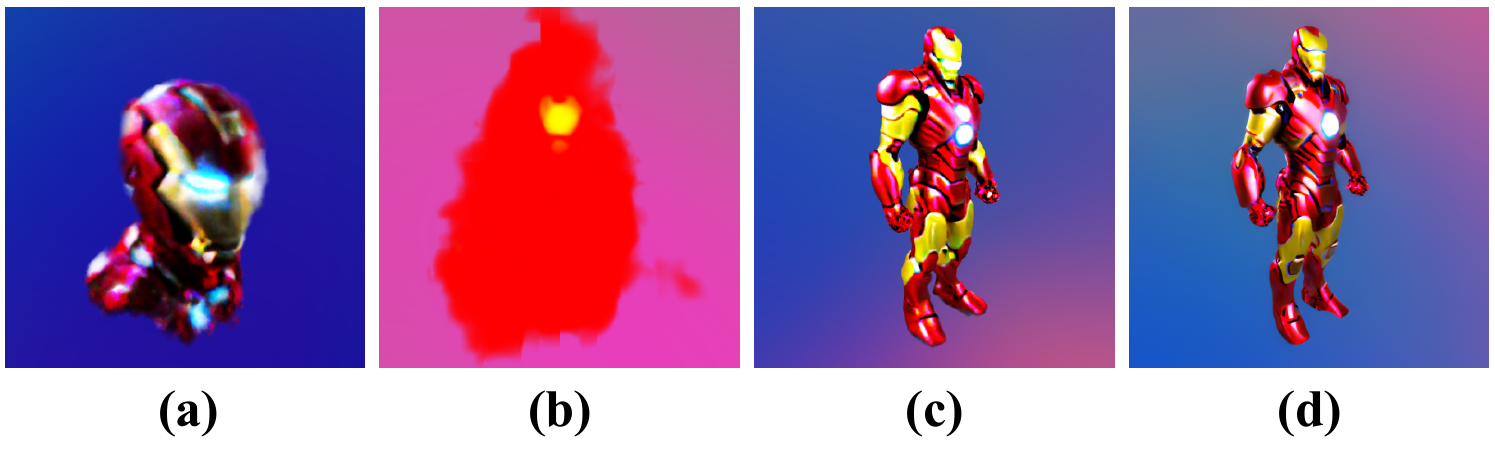}
    \caption{Ablation study. Fig(a) is without the initialization stage. Fig(b) is without the boost stage. Fig(c) is without the self-boost stage. Fig(d) is our complete BoostDream method.}
    \label{fig:figure5}
\end{figure}

In the proposed pipeline, we delineate three distinct stages: the initialization stage, the boost stage under the guidance of a coarse 3D asset, and the self-boost stage under the guidance of its intrinsic multiview. To elucidate the impact of each stage, we undertake an ablation study where we systematically eliminate each stage in isolation to assess their contributions. The results are shown in Figure \ref{fig:figure5}, which illustrates that instead of using the initialization stage, the model may not converge quickly and have wrong guidance from the 3D assets we want when randomly initialized. Without coarse 3D asset guidance, the model demonstrates a propensity to evolve towards unforeseen outcomes. This phenomenon can be attributed to the premise that post-initialization errors perpetuate a cycle of learning based on these inaccuracies. Finally, without its self-guidance, the model is constrained by the original coarse 3D asset, thus failing to achieve a level of detail and quality that is otherwise attainable. 

We also conducted two other ablation studies. One uses the coarse 3D asset generated by the feed-forward method to initialize the traditional SDS-based method. Another is using different control conditions to replace normal maps, such as canny edge and depth maps. See more ablation studies in the Appendix \cite{Boost2024}.

\subsection{User Study}\label{sec:User}
\begin{figure}[ht]
    \centering
    \includegraphics[width = 8cm]{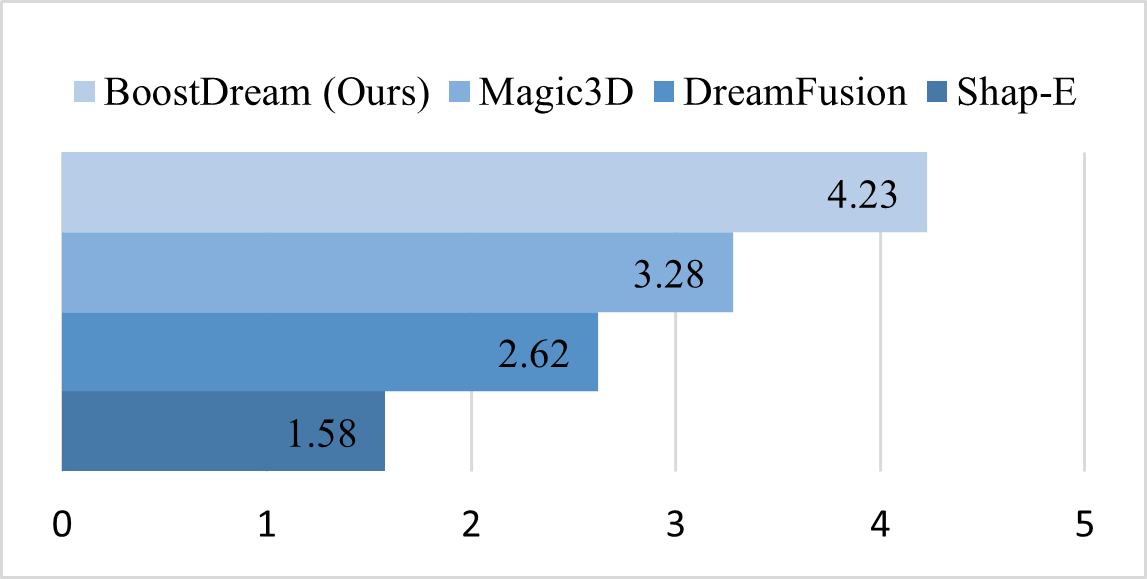}
    \caption{User study. Our model demonstrates a significant superiority in user preference, achieving high scores in a blinded survey with a maximum rating of 5. Here, we use BoostDream-NeRF as our implementation.}
    \label{fig:figure6}
\end{figure}

Due to the lack of existing evaluation metrics for 3D generation, we conduct a user study to assess model performance. We create an evaluation set generated through 30 prompts using four methods. For a specific 3D asset, a rendered video with an input text prompt was shown to participants. 20 participants are invited to individually score each item in the set, focusing on prompt relevance and the quality of generated details, with scores ranging from 1 to 5. As shown in Figure \ref{fig:figure6}, the results demonstrate that our method is markedly superior.

\section{Conclusion}
In summary, our research introduces BoostDream, a novel method that seamlessly combines differentiable rendering and text-to-image advancements to generate high-quality 3D assets efficiently. Central to our approach is the multi-view SDS, which effectively overcomes the Janus problem in generation processes. 
BoostDream can be applied to various differentiable 3D representations, generally improving the quality and reducing the time consumption of existing 3D generation methods.
This work not only advances the quality and diversity of 3D assets but also sets a precedent for future innovations in 3D modeling, with wide-reaching implications for virtual reality and gaming industries.

\newpage

\section*{Ethical Statement}

The BoostDream proposed in this paper uses the 2D diffusion model as the prior to generating high-quality 3D assets with a multi-view strategy. We aim to reduce the cost of creating highly detailed 3D assets. However, we do note that the method could be applied to generate disinformation or violent and sexual content. With 2D diffusion as prior, it also inherits similar ethical and legal considerations to problematic biases and limitations that 2D diffusion models may have. 


\bibliographystyle{named}
\bibliography{ijcai24}

\appendix
\input{ijcai24_appendix}

\end{document}

%% file: ijcai24_appendix.tex
\section*{Appendix}
\appendix

\section{Implementation Details}
In the training process of the NeRF \cite{mildenhall2021nerf} setting, we set rendering resolution to $128 \times 128$, and batch size to 1. We apply our random multi-view render system to capture a combined image with four sub-images with rotation angle $\alpha$ set to $90^\circ$. We use AdamW optimizer \cite{kingma2014adam} with learning rate $1\times10^{-2}$ and $1\times10^{-3}$ for geometry and background modeling. The background is replaced with random colors with $80\%$ of chance. In the DMTet \cite{shen2021dmtet} setting, most of the parameters stay the same, but in the self-boost stage, we increase the resolution to $512 \times 512$
 for a better result. The initialization stage of 3D Gaussian Splatting \cite{kerbl3Dgaussians} is somehow different from the other two methods as they use hash-grid while 3D Gaussian Splatting is able to initialize from point cloud representation directly. The rendering resolution is also $512 \times 512$.
 
 We apply the CFG trick and negative prompts following the example from MVDream \cite{shi2023mvdream}, further append prompt ``, 3d asset” or ``, multi-view of the 3d asset" to get a more consistent result.

\begin{figure*}[!h]
    \centering
    \includegraphics[width=\textwidth]{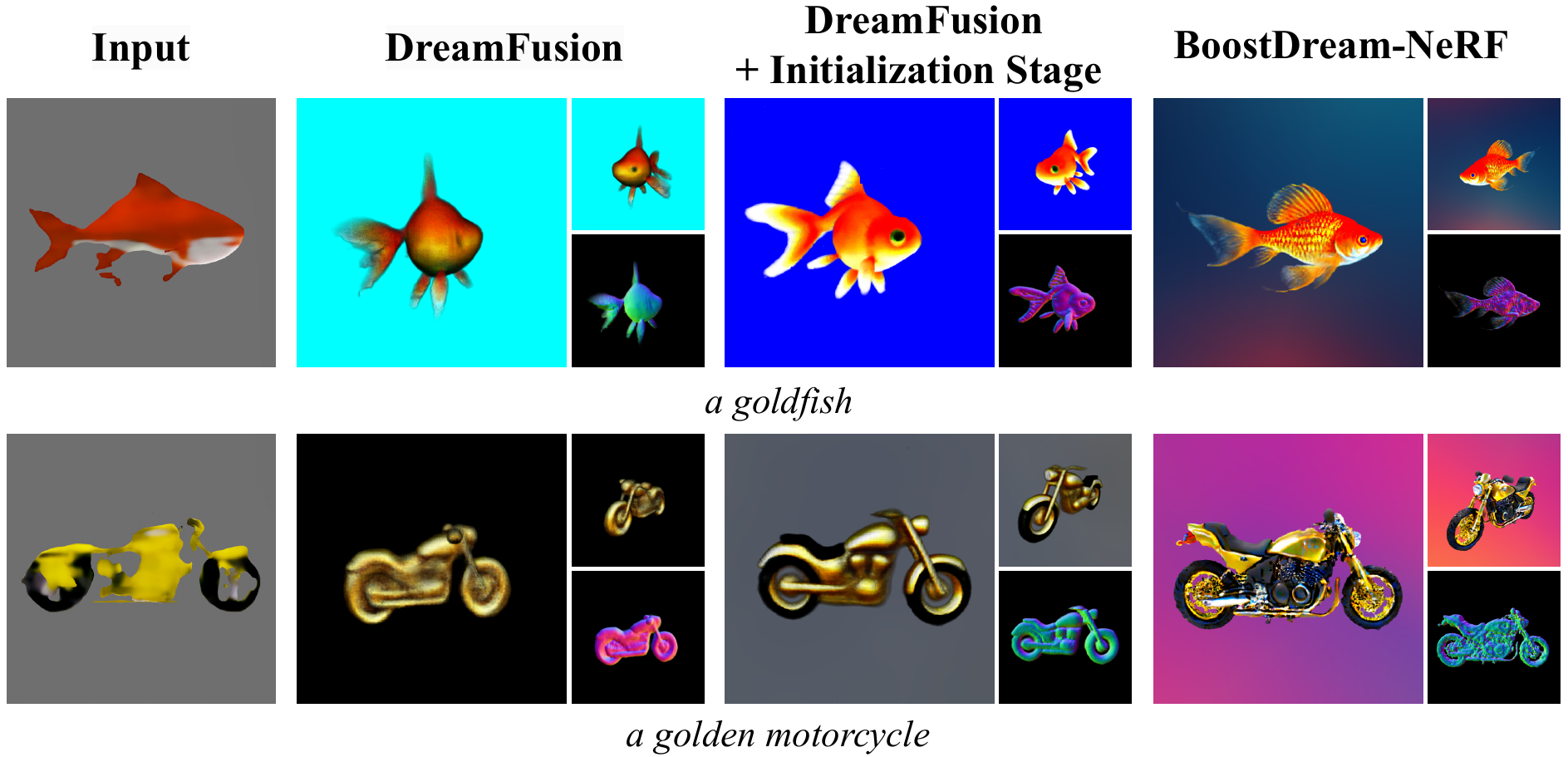}
    \caption{Simply Combination Ablation Study. The first column is the input coarse model generated by Shap-E \protect\cite{jun2023shape}, while the next three columns are the results for the original DreamFusion \protect\cite{poole2022dreamfusion}, the DreamFusion with initialization stage, and our BoostDream-NeRF, respectively.}
    \label{fig:figure1A}
\end{figure*}

\section{Simply Combination Ablation Study}
Our BoostDream method does not just simply combine the feed-forward approach with
the SDS-based method. To further test the benefits of applying our multi-view based strategy, we also design an ablation study using DreamFusion \cite{poole2022dreamfusion} with the same initialization stage as our method. We use the results from Shap-E \cite{jun2023shape} in the initialization stage and use the same prompt text as input to optimize the NeRF representation with DeepFloyd \cite{Deepfloyd}. The results of the original DreamFusion, the DreamFusion with initialization stage, and our BoostDream-NeRF are shown in Figure \ref{fig:figure1A}. We can see in the first row even with the proper initialization, DreamFusion still suffers from the Janus problem and has coarse results compared to our BoostDream results.

\section{Control Condition Ablation Study}
\begin{figure*}[htbp]
    \centering
    \includegraphics[width=\textwidth]{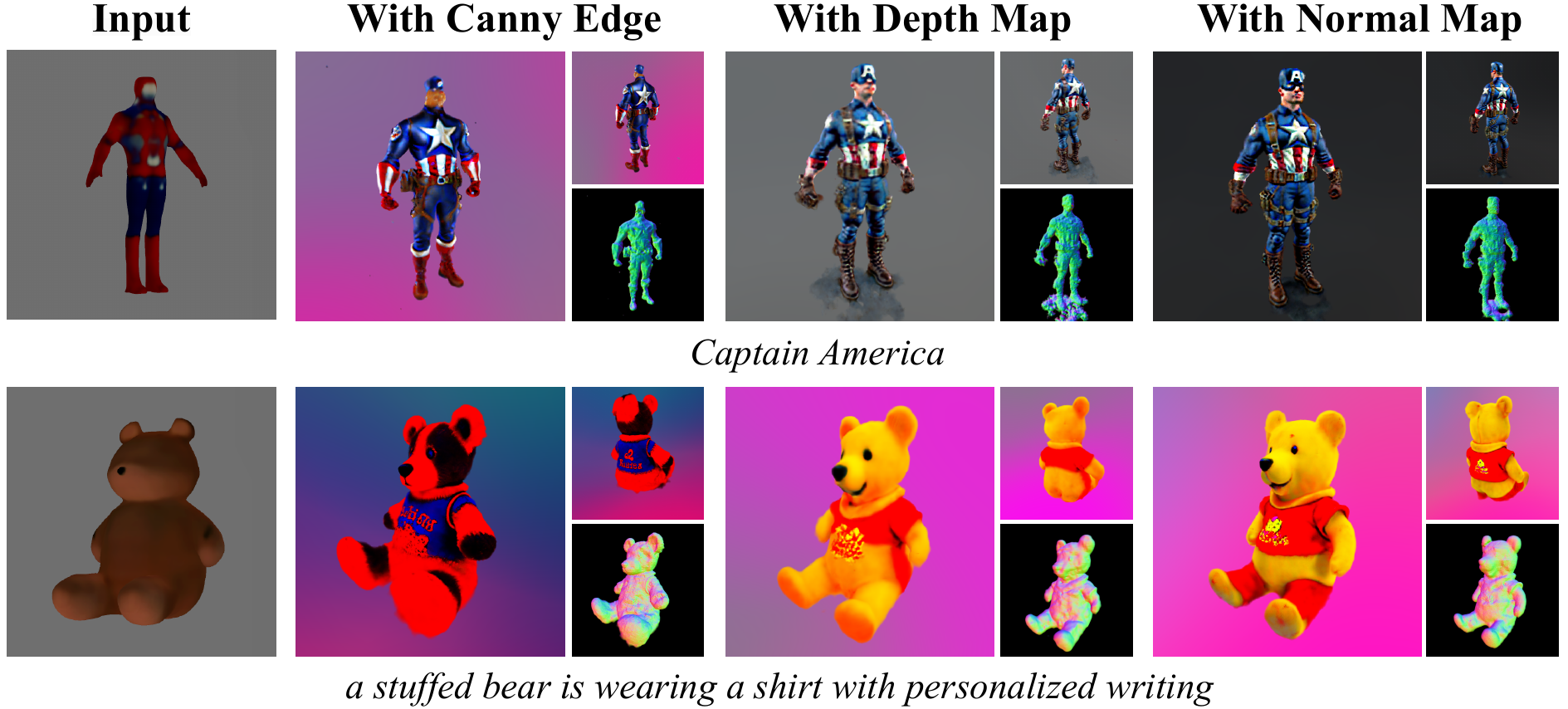}
    \caption{Control Condition Ablation Study. The first column is the input coarse model generated by Shap-E \protect\cite{jun2023shape}, while all other columns are the output of our BoostDream method with different control conditions.}
    \label{fig:figure2A}
\end{figure*}

We also test our method with different multi-view control conditions replacing the normal map. We choose canny edge \cite{canny1986computational} and depth map \cite{ranftl2020towards} as guidance obtained through the same multi-view render system as normal map. The results are shown in the Figure \ref{fig:figure2A}. Canny edge just contains the edge information of the 3D asset. Intuitively, it is unsuitable as a control condition when generating high-quality 3D assets. The results also illustrate this point: when using canny edge as the control condition, the 3D asset suffers from incomplete generation. Especially in the second row, the bear turns out to be unnatural and has strange colors. Instead of canny edge using edge information to guide the refinement process, the depth map utilizes depth information, leading to complete generation results. However, we find that the generated results are less detailed when the control condition is depth map. This can be explained by the fact that minor details information is not prominent in depth map but salient in normal map \cite{zhang2023adding}. We can further validate this idea with the last column, the generated 3D assets are high-quality and with more details when under the guidance of normal map.

\section{Result on Different 3D Representations}
This section supplements the comparison experiment in Section 4.3. We implement our BoostDream on other differentiable representations, including DMTet \cite{shen2021dmtet} and 3D Gaussian Splatting \cite{kerbl3Dgaussians}.
The results are shown in Figure \ref{fig:figure3A}, illustrating the generality of our method in generating high-quality assets using different differential 3D representations.
\begin{figure*}[t]
    \centering
    \includegraphics[width=\textwidth]{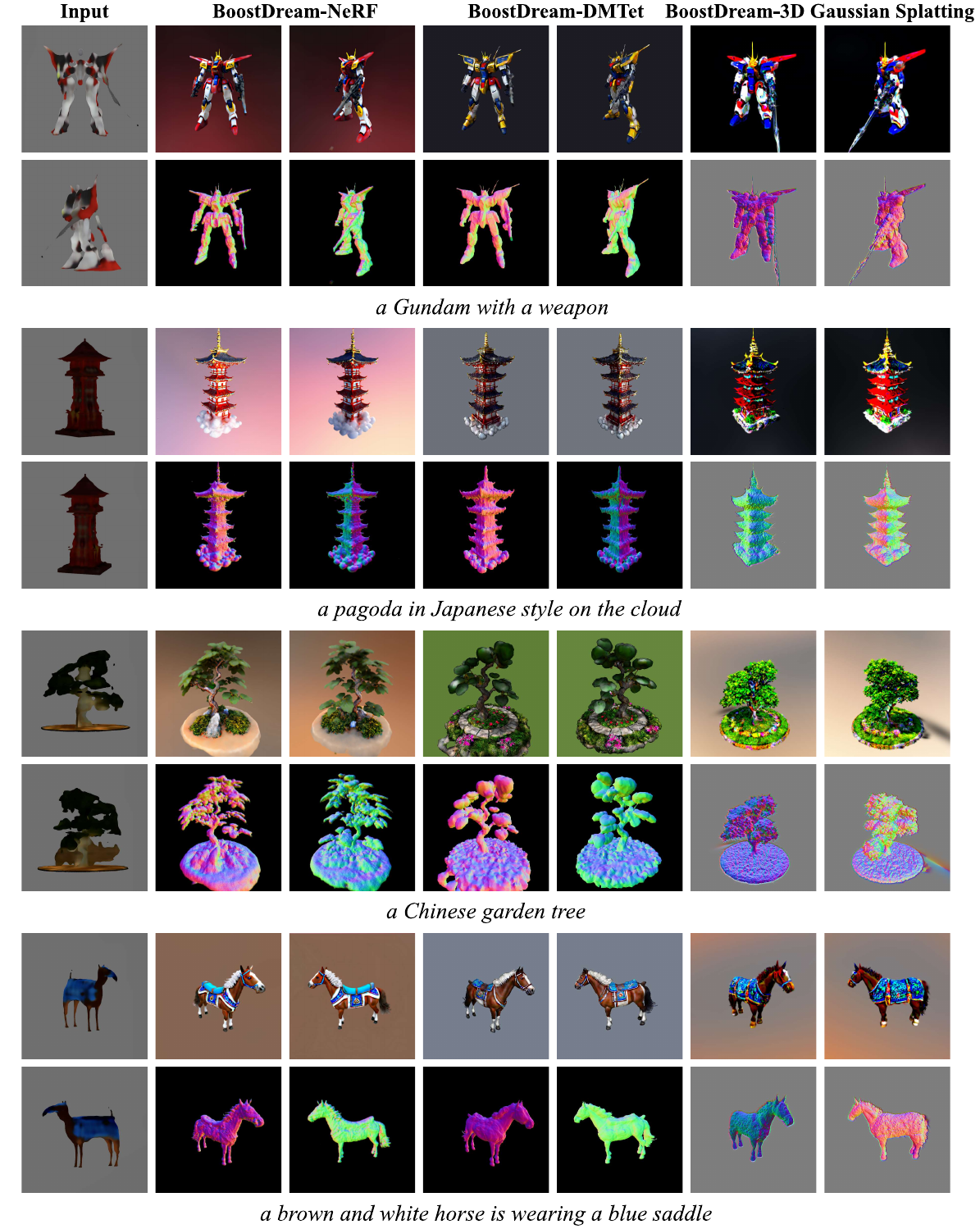}

    \label{fig:figure3A}
\end{figure*}

\begin{figure*}[t]
    \centering
    \includegraphics[width=\textwidth]{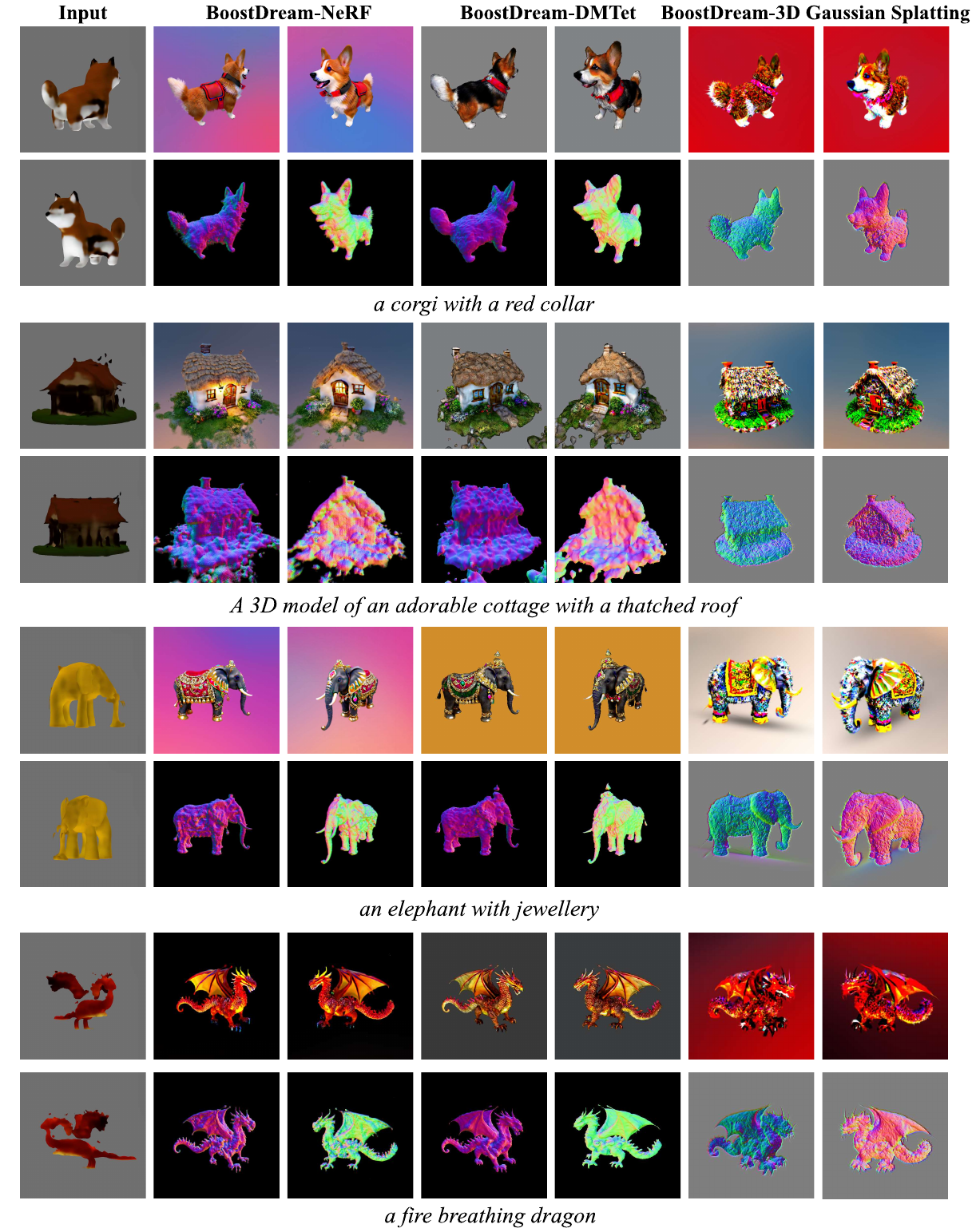}
    \caption{Result on Different 3D Representations.The first column is the input coarse model generated by Shap-E \protect\cite{jun2023shape}, while the next three columns are the results of our BoostDream method implemented with NeRF \protect\cite{mildenhall2021nerf}, DMTet \protect\cite{shen2021dmtet} and 3D Gaussian Splatting \protect\cite{kerbl3Dgaussians}, respectively. }
    \label{fig:figure3A}
\end{figure*}